\documentclass{article} 

\usepackage{fullpage}
\usepackage{microtype}
\usepackage{graphicx}
\usepackage{subfigure}
\usepackage{booktabs} 
\usepackage{hyperref}

\RequirePackage{natbib}

\usepackage[utf8]{inputenc}

\usepackage{xcolor}
\hypersetup{
    colorlinks=true,
    linkcolor=blue,
    filecolor=magenta,      
    urlcolor=cyan,
    citecolor=blue,
}

\usepackage{algorithm}
\usepackage{algorithmic}

\usepackage{amsmath}
\usepackage{amsthm}
\usepackage{dsfont}
\usepackage{amssymb}
\usepackage{amscd}
\usepackage{mathtools}
\mathtoolsset{showonlyrefs=true}
\usepackage{enumerate}
\usepackage{authblk}
\usepackage{ bbold }
\usepackage{wrapfig}
\usepackage[T1]{fontenc}
\usepackage{cleveref}

\usepackage{xcolor}
\definecolor{darkgreen}{rgb}{0.0, 0.5, 0.0}

\newtheorem{theorem}{Theorem}

\newcommand{\argmax}{\operatorname*{argmax}}

\newcommand{\kl}[2]{\operatorname*{KL}(#1||#2)}

\newcommand{\R}{\mathbb{R}}

\newcommand{\uc}{\mathcal{U}}
\newcommand{\states}{\mathcal{S}}
\newcommand{\actions}{\mathcal{A}}
\newcommand{\un}{\mathbb{1}}
\newcommand{\E}{\mathbb{E}}

\newcommand*\samethanks[1][\value{footnote}]{\footnotemark[#1]}

\title{On the importance of data collection for training general goal-reaching policies}

\author[1]{Alexis Jacq\thanks{Equal contribution.}}
\author[1]{Manu Orsini\samethanks}
\author[1]{Gabriel Dulac-Arnold}
\author[1]{Olivier Pietquin}
\author[1]{Matthieu Geist}
\author[1]{Olivier Bachem}

\affil[1]{Google Research, Brain Team}

\date{} 

\begin{document}

\maketitle

\begin{abstract}
Recent advances in ML suggest that the quantity of data available to a model is one of the primary bottlenecks to high performance.  Although for language-based tasks there exist almost unlimited amounts of reasonably coherent data to train from, this is generally not the case for Reinforcement Learning, especially when dealing with a novel environment.  In effect, even a relatively trivial continuous environment has an almost limitless number of states, but simply sampling random states and actions will likely not provide transitions that are interesting or useful for any potential downstream task.  \textit{How should one generate massive amounts of useful data given only an MDP with no indication of downstream tasks? Are the quantity and quality of data truly transformative to the performance of a general controller?}  We propose to answer both of these questions.  First, we introduce a principled unsupervised exploration method, ChronoGEM, which aims to achieve uniform coverage over the manifold of achievable states, which we believe is the most reasonable goal given no prior task information.  Secondly, we investigate the effects of both data quantity and data quality on the training of a downstream goal-achievement policy, and show that both large quantities and high-quality of data are essential to train a general controller: a high-precision pose-achievement policy capable of attaining a large number of poses over numerous continuous control embodiments including humanoid.
\end{abstract}

\section{Introduction}
Recent work in large language models~\citep{Palm}, as well as general conclusions such as though proposed by~\citep{sutton2019bitter} suggest that one of the main bottlenecks of current machine-learning methods is access to large amounts of informative data.  This is particularly true for the domain of Reinforcement Learning (RL)~\citep{sutton2018reinforcement}, where data acquisition can be costly, and coherent datasets~\citep{gulcehre2020rl, fu2020d4rl} are less prominent and smaller than text datasets pulled from the internet.  
\\
This brings us to the two core questions of this paper: 
\\

\textit{How should one generate massive amounts of useful data given only an MDP with no indication of downstream tasks?}

\textit{What are the effects of quantity and quality of data on the training of a general policy?}
\\
\\
To answer the first question, we present Chronological Greedy Entropy Maximisation (ChronoGEM), a principled, highly scalable exploration method whose goal is to achieve uniform coverage over the manifold of achievable states. We demonstrate both theoretically that ChronoGEM approximates uniform sampling, and empirically that states achieved by ChronoGEM are diverse and well-distributed.
\\
Regarding the second question, we investigate the effects of data quality and quantity using Entropy-Based Conditioned Continuous Control Policy Optimization (C3PO), a goal-conditioned policy training pipeline that leverages pre-defined datasets of uniform goal states to learn a general pose-attainment policy on various high-DoF control tasks.  Data quality is analyzed by comparing performance of our policy trained with goals sampled from ChronoGEM vs other contemporary unsupervised exploration methods. The question of quantity is particularly interesting if we can analyze the asymptotic effects of data, perhaps far beyond what is generally realized in RL training.  The main blocker to this type of experiment in the past has simply been the time such experiments would take to run, but given observations that the long-tail of training can still provide significant performance benefits~\citep{silver2017mastering}, it is important to understand how much is being left off the table by not pushing this limit.  Recent advances in accelerator-specific parallelizable physics simulators such as Brax~\citep{brax2021github} can allow us to train our policies into the billions of steps in a matter of hours, thus allowing us both to tune said approaches and observe their asymptotic behavior when pushing training orders of magnitude beyond what is usually performed.
\\
\begin{figure*}[t]
    \centering
    \includegraphics[width=\linewidth]{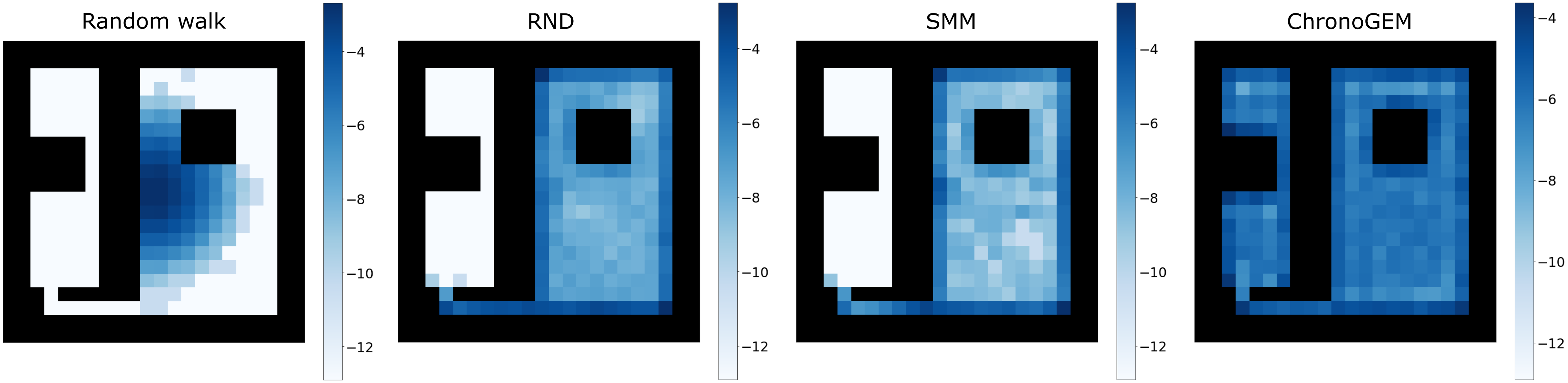}
    \vspace{-20pt}
    \caption{Log-frequencies of the discretized states visitation when taking the last states from 4000 episodes, sampled according to (left to right) a random walk, SMM, RND and ChronoGEM, averaged over 10 seeds. Only ChronoGEM managed to visit states in the top-left room.}
    \label{fig:maze_exploration}
    \vspace{-10pt}
\end{figure*}
\\
We structure our paper by first introducing our two methods in Section \ref{sec:methods}: We look at both the theoretical foundations of ChronoGEM, as well as its independent building blocks.  We also introduce the C3PO training setup, and discuss how it can be used to evaluate the effects of quantity and quality of data for learning a generalist policy.  In Section \ref{sec:exp} we perform a series of experiments to attempt to answer our core questions.  We begin by comparing the performance of ChronoGEM relative to existing unsupervised exploration methods such as RND~\cite{burda2018exploration} and SMM~\citep{lee2019efficient} on a series of environments, ranging from an illustrative 2-D maze task, to high-DoF continuous control environments through various methods.  We then investigate the effects of quantity and quality of data on the training of a general controller policy.  We do this by  comparing goal achievement across our different goal datasets, generated by either ChronoGEM, RND or SMM, and show that policies learnt using ChronoGEM data can achieve a much more diverse set of goals than those learnt on other data sources. Finally, we push the training regime to multiple billions of steps to examine the asymptotic performance of our policy architecture, and achieve high-precision pose-attainment even on a complex system such as Humanoid. Finally, we briefly demonstrate applicability of a general controller in a zero-shot imitation task. Section \ref{sec:related} situates our proposed methods relative to current research.

\section{Methods: ChronoGEM \& C3PO}
\label{sec:methods}
We introduce our two contributions:  Chronological Greedy Entropy Maximisation (ChronoGEM) \& Entropy-Based Conditioned Continuous Control Policy Optimization (C3PO).  ChronoGEM is our proposed solution for principle scalable unsupervised exploration of high-DoF environments which we use to investigate how to generate massive amounts of useful data for downstream RL tasks.  C3PO is our proposed pose-attainement policy training method to investigate the effects of quantity and quality of data on the performance of a generalist policy.
\\
To clarify notations we will quickly introduce the Markov Decision Process (MDP), which can be defined by a transition function $T$ which maps states $s \in \states$ and actions $a \in \actions$ to a corresponding state $s_{t+1} = T(s_t, a_t)$, while potentially providing a scalar reward $r(s_t, a_t) \in \mathbb{R}$ if there is an associated task.
\looseness=-1

\subsection{Chronological Greedy Entropy Maximisation (ChronoGEM)}

Given an arbitrary Markov Decision Process (MDP), and the goal of generating massive amounts of useful data for arbitrary downstream tasks, and no prior to guide our exploration of the MDP, we argue that the ideal set of explored states should be uniformly sampled from the manifold of reachable states at a given horizon $T$.  The shape of the manifold of reachable states is cannot be known \textit{a priori}, and may be arbitrarily complex, thus rendering any form of direct sampling impossible. 
Given acces to the MDP, this sampling can however be approximated by an iterative algorithm.
\looseness=-1
\\
Let us assume we have a sample of $N$ states that are approximately uniform for a horizon of $T-1$.  We can perform $K$ uniform actions from each of these states, and result with $NK$ states, whose distribution will be biased by the MDPs natural dynamics instead of being uniform across state values.  Nevertheless, with a sufficiently large $N$ there is likely some subset of these $NK$ states which would can be sampled to approximate a uniform distribution over states at horizon $T$.  
Let $\rho_T$ be the distribution induced by these $NK$ next states, and let us assume we have a method to estimate $\rho_T$.
Since the set of achievable states in $T$ steps is generally bounded, and given that we are able to closely estimate $\rho_T$, we can sub-sample the $NK$ states with probability $\frac{1}{\rho_T}$, which will approximate uniform sampling according to the state statistics maintained in $\rho_T$. We prove in Appendix~\ref{appendix:uniform_sampling} that such sub-sampling approximates uniform sampling when the number of sampled states is sufficiently large. 
\\
Given this recursive definition, we can begin with states sampled from the environment's initial starting-state distribution $\rho_0$, perform $K$ uniform actions and generate $NK$ states. We can then sub-sample the $NK$ states back to $N$ states that approximate the uniform distribution using inverse probability weighting according to our density estimator $\rho$.  By iterating this process $T$ times, we can obtain a set of $N$ states that are uniform for a given horizon $T$.
We call this process ChronoGEM (for Chronological Greedy Entropy Maximization) since at a given step, it only focuses on maximizing the entropy by directly approximating a uniform distribution over the next step, without further planning. ChronoGEM is described more formally in Algorithm~\ref{alg:chronogem}. 
\begin{algorithm}
     \caption{ChronoGEM}
     \label{alg:chronogem}
    \begin{algorithmic}[1]
    \STATE Sample $N$ states $S_0=\{s_0^i\}_{i=1}^N \sim \rho_0$.
    \FOR{$t=1$ {\bfseries to} $T$}
        \STATE Sample $K$ uniform actions for each state of $S_{t-1}$.
        \STATE Obtain $KN$ next states.
        \STATE Estimate $\rho_t$ using a density model fitted on the distribution of these $KN$ states.
        \STATE Sample $N$ states with  $p(s)\propto \frac{1}{ \rho_t(s)}$ to get $S_t$.
    \ENDFOR
    \STATE Return $S_T$.
    \end{algorithmic}
\end{algorithm}

ChronoGEM requires exactly $KNT$ interactions with the environment, which makes it easily controllable in term of sample complexity.  Due to the $N$ sampled states being independent, ChronoGEM can be parallelized with $N$ jobs consuming $KT$ interactions each, significantly shortening the time complexity.  As implemented, ChronoGEM requires re-settable states, although this could potentially be relaxed with a return-to-state policies similar to `First Return then Explore'~\citep{ecoffet2021first}, refer to \Cref{app:resettable} for more discussion. 

\looseness=-1

\paragraph{Density estimation.}
\label{density_estimation}
As we saw above, ChronoGEM requires a density estimator at each iteration to estimate $\rho_t$.    
Many choices of density estimation in high dimensional space exist, from simple Gaussian estimators to neural network-based methods such as autoregressive models or normalizing flows. 
The performance of these models may vary given the type of data: Some models are more suited for images, while others are better for text or lower dimensions. 
We implemented 7 candidate models, including Gaussian models (Multivariate, Mixture), autoregressive networks (RNade~\citep{uria2013rnade}, Made~\citep{germain2015made}), and normalizing flows (real-NVP~\citep{dinh2016density}, Maf~\citep{papamakarios2017masked}, NSF~\citep{durkan2019neural}).
After comparing performance of the various models through a state modeling task decribed in \Cref{appendix:density}, we concluded that the NSF variant of normalizing flows worked best for the state modeling task.
\looseness=-1

\subsection{Entropy-Based Conditioned Continuous Control Policy Optimization (C3PO)}
\label{sec:c3po}
C3PO's objective is to learn a generalist control policy that can attain any reachable state in a given environment.  Our intent is to investigate the effects of data quantity and quality on C3PO's training regime, and better understand the importance of both diverse data, and the asymptotic effects of massive amounts of experience on policy quality.
\\
C3PO is a training regime that wraps an arbitrary policy learning algorithm with a curriculum and success threshold annealement.  It requires a dataset of goal states $s_g \in \mathcal{G}$, as well as a goal-achievement criteria $S: (\states, \states, \mathbb{R}) \rightarrow \{0,1\}$.  The overall training regime is defined in \Cref{alg:c3po}. 
In practice this training loop is run in a highly-parallelized manner on a set of accelerators, allowing for significantly longer episodes than is usually performed in RL experiments.  The success criteria's threshold is initialized to a relatively tolerant value, and conditionally annealed every time the learnt policy achieves $90\%$ success rate according to the goal-achievement criteria.
\\
As we will see in \Cref{sec:exp_general_quality}, by using various sources of goal states, such as ChronoGEM, RND, SMM, and random walk for which we will have already compared the state space coverage and relative entropy, we can understand the downstream effects of data quality on training a general policy.  With regards to data quantity, we will see in \Cref{sec:exp_general_quantity} that by pushing the number of training steps into the billions, we can significantly improve goal achievement rates of the generalist policy.
\looseness=-1

\begin{algorithm}[H]
     \caption{C3PO}
     \label{alg:c3po}
    \begin{algorithmic}[1]
    \REQUIRE Goals $\mathcal{G}$, Achievement Criteria $S$, Threshold $\epsilon$
    \STATE $\texttt{success\_rate} = 0$, $\mathcal{D} = \emptyset$
    \WHILE{True}
        \STATE Draw goal $s_g \in \mathcal{G}$
        \STATE Rollout $\pi(\cdot, s_g)$, until $S(s_t, s_g, \epsilon) == 1$ or $t == T$
        \STATE Add rollout to $\mathcal{D}$
        \STATE Update \texttt{success\_rate} average with $S(s_t, s_g, \epsilon)$
        \STATE \texttt{IF} ($\texttt{success\_rate} > 0.9): \epsilon =  0.99 \times \epsilon$
        \STATE Run policy training algorithm on $\mathcal{D}$.
    \ENDWHILE
    \STATE \textbf{return} $\pi$
    \end{algorithmic}
\end{algorithm}

\section{Experiments}
\label{sec:exp}
We will now detail the experiments performed to investigate our two core questions. First we will look at experiments on \textit{how one should generate massive amounts of useful data}, by investigating the performance of ChronoGEM relative to other unsupervised exploration algorithms.  We will then look at the effects of \textit{data quantity and quality} by looking at C3PO performance as these two factors are varied.

\subsection{Environments}
\paragraph{2D Maze.}
As a tool scenario to test ChronoGEM, we implemented a two-dimensional continuous maze in which actions are $d_x,d_y$ steps bounded by the $[-1, 1]^2$ square, and the whole state space is bounded by the $[-100, 100]^2$ square. This environment, illustrated in Fig?~\ref{fig:maze_exploration} is similar to the maze used by~\citet{kamienny2021direct}, except it adds a significant difficulty induced by the presence of two narrow corridors that needs to be crossed in order to reach the top-left room.

\paragraph{Continuous control tasks.}
\label{continuous_control}
We use control tasks from Brax~\citep{brax2021github} as high dimensional environments.  We chose four environments to cover varying complexities of task: Hopper, Walker2d, Halfcheetah and Ant.
Environment observations were modified to contain $(x, y, z)$ positions of all body parts to be better aligned with pose-achievement goals.
All measures (cross-entropy in Section~\ref{entropy_exp} and reaching distances in Section~\ref{goal_training}) are based on this observation space.
To attenuate energy accumulation corner cases in the simulator, we maintain the environments in a low energy regime by reducing the action amplitude by a factor of $0.1$ for Hopper and Walker, $0.01$ for HalfCheetah. Actions for Ant are unmodified.
In the two following subsections, we considered episodes of length $T=128$. To compensate for varying control frequencies while keeping the same wall clock episode length, there is an action repeat of $6$ for Hopper and Walker.
All default episode end conditions (primarily involving falling) have been removed.

\subsection{Generating Useful Data With ChronoGEM}
In our first set of experiments, we look at the performance of ChronoGEM relative to three other unsupervised exploration algorithm: RND~\citep{burda2018exploration}, SMM~\citep{lee2019efficient} and a random walk policy.  In \Cref{sec:exp_chrono_maze} We will begin by looking at an easily interpretable environment, a 2D maze.  We will then look at more complex environments and quantify data coverage in \Cref{sec:exp_chrono_entropy}. 
ChronoGEM was run with $N=2^{17}$ parallel environments and branching factor $K=4$ in all experiments, except for Humanoid where $N=2^{15}$ and $K=64$ (Humanoid required a larger branching factor to avoid absorbing states).  ChronoGEM-related hyperparameters are in \Cref{tab:chrono-hypers} (Appx.).

\paragraph{Chronogem Visualization: 2D Maze.}
\label{sec:exp_chrono_maze}

The main goal of this experiment is to verify that ChronoGEM  manages to induce a uniform distribution over the whole state space of a simple yet challenging toy environment. In order to emphasize the relative difficulty of the exploration of this maze, we also run SMM, RND and a random walk to compare the resulting state coverage. 
In this setup, we know that if $T$ is large enough, all achievable states are just every point in the maze, so ChronoGEM should be uniform on the maze given sufficient time (for instance, $T=1000$). We can see the final state distribution in Figure~\ref{fig:maze_exploration}, and observe that while ChronoGEM achieves fairly uniform state coverage, both RND and SMM fail at exploring beyond the first corridor, and a random walk did not even explore the whole first room.  This suggests that ChronoGEM is performing as expected, and that RND and SMM struggle with bottleneck states, whereas random walk has trouble with diffusing far from the starting state distribution.

\begin{figure*}[t]
    \centering
    \includegraphics[width=\linewidth]{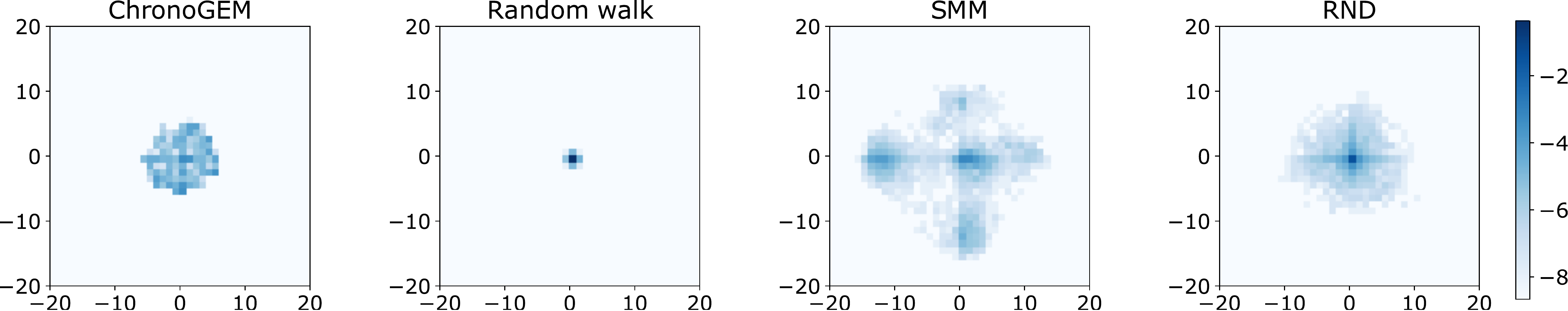}
    \vspace{-15pt}
    \caption{Sky-view of the discretised spatial log-frequency covered by Ant with the different exploration methods. SMM has the largest scope but contains empty zones even close to the origin. Both RND and ChronoGEM share similar range covering all directions, and ChronoGEM is visibly much more uniform, while other methods are concentrated on the origin. Note that this only represents the spatial positions, while both poses and positions are being explored.}
    \label{fig:helicopter}
\end{figure*}

\begin{figure*}[t]
    \centering
    \includegraphics[width=\linewidth]{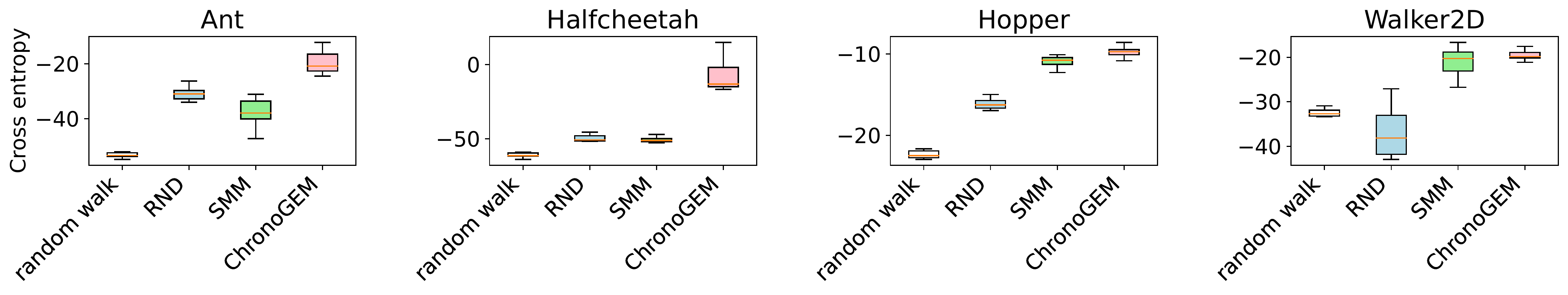}
    \vspace{-20pt}
    \caption{Distribution over 10 seeds of the cross entropies of the state visitation induced by ChronoGEM, RND, SMM and a random walk, on different continuous control tasks.}
    \label{fig:entropy}
    \vspace{-10pt}
\end{figure*}

\paragraph{State Coverage Quantification with Entropy.}
\label{entropy_exp}
\label{sec:exp_chrono_entropy}
In this section we will look at a more quantified method for evaluating state coverage of an exploration method using our learnt entropy estimator.  
Given a set of points $x_1\hdots x_N$ sampled from a distribution with an unknown density $\rho$, one can estimate an upper-bound of the entropy of $\rho$ given by the cross-entropy $H(\rho, \hat{\rho})$ where $\hat{\rho}$ is an estimation of $\rho$:
\begin{equation}
    H(\rho, \hat{\rho}) = -\E_{x\sim \rho}[\log \hat{\rho}(x)] 
    = H(\rho) + \kl{\rho}{\hat{\rho}} \geq H(\rho).
\end{equation}
The estimate $\hat{\rho}$ being learned by maximum likelihood specifically on the set of points, it directly minimises the cross entropy and closely approximates the true entropy.  
The KL term becomes negligible and only depends on the accuracy of the trained model on the observed set of points, which supposedly does not differ given the different exploration method that generated the points. Consequently, comparing the cross-entropy is similar to comparing the entropy of the distribution induced by the exploration.
In this experiment, we used this upper-bound to study the efficiency of ChronoGEM compared to RND, SMM and a random walks. Figure~\ref{fig:entropy} displays box plots over 10 seeds of the resulting cross entropy measured on the sets of states induced by different algorithms, on the 4 continuous control tasks. As expected, the random walk has the lowest entropy, and SMM has a better entropy than RND on Hopper and Walker2d (which makes sense since it is optimizing for the maximization of the entropy). ChronoGEM has the highest entropy on all environments, especially on HalfCheetah, where it was the only method to manage exploration while the actions were drastically reduced by the low multiplier (see Section~\ref{continuous_control}).
In order to illustrate the fact that ChronoGEM induces a state distribution that is close to uniform, we measure the spatial coverage based on a discrete grid of the x-y plan: if the distribution is uniform over both the possible poses and positions, it should be in particular uniform over the positions.
Figure~\ref{fig:helicopter} shows the resulting log-frequency on the x-y grid visitations and although ChronoGEM does not have the biggest exploration horizon, it nevertheless provides the most uniform coverage. Grid visitation for Hopper, Walker2d and Halfcheetah are available in \Cref{fig:visits} in \Cref{appendix:visitation}.
Given both the entropy coverage and the various qualitative visualizations, we believe ChronoGEM provides consistently uniform and exhaustive coverage of the state space for a given horizon $T$.  Additionally, it is an efficient algorithm, which can quickly generate large amounts of data.  For these reasons we believe it is a good candidate for generating massive amounts of useful data for downstream tasks.  We will now look at this in practice by comparing its utility relative to RND, SMM and randomw-walk datasets as goal states for training a general goal-achievement policy.

\subsection{Effects of Data Quality and Quantity on General Policy Training with C3PO}
\label{goal_training}
\label{sec:exp_general_quality}
In this section we investigate the effects of data quality and quantity on the training of a general policy using C3PO.  We start by looking at data quality, by evaluating C3PO policies trained on varied datasets of goal states.  We hope to better understand the importance of data coverage for the performance of training a general goal-attainment policy.  
Indeed, the ideal goal distribution would be uniform across all achievable states, so if the data was generated in a way that it best approximates this distribution, we would hope this would significantly impact the quality of our trained policy. 
For each environment, we run the four exploration methods (ChronoGEM, Random Walk, SMM and RND) with 3 seeds each, which we split into training \& evaluation goal sets. Training goal sets have 4096 goals and evaluation goal sets have 128 goals.  For the task of goal-achievement, we use the following reward: $r(s_t, g) = -\|s_t - g \|_{\infty}^2 = -\max_{\mathbf{b_i} \in \mathbf{s_t}}\|\mathbf{b_i} - \mathbf{g_i} \|^2$
, where $\mathbf{b_i}$ is a set of coordinates corresponding to a body component of the embodiment (leg, torso, foot).  This reward effectively penalizes according to the distance of the most distant body part relative to the desired goal pose.  The success criteria is $S(s_t, g, \epsilon) = |r(s_t, g)| <\epsilon$, thus effectively ending the episode once a certain value of reward is attained.  The underlying training algorithm used is SAC~\citep{haarnoja2018soft} with the hypers detailed in ~\Cref{tab:c3po-hypers}.  Goal-states are simply appended to the observation vector of the policy network.

\paragraph{Effects of Data Quality.}
\looseness=-1
We evaluate policy performance with two quantitative and one qualitative approach.  We perform cross-validation across datasets, by evaluating each policy trained with one dataset on evaluation goals from all other datasets.  We observe that C3PO trained on ChronoGEM data is the most robust across datasets and environments, matching or often beating policies evaluated on their own datasets, especially for low tolerance values of $\epsilon$.  This can be explained by the fact that C3PO learns to reach a high variety of poses, since being able to achieve poses with high fidelity is what matters for low distance threshold regime.  We believe that slightly lower performance on some environment/dataset pairs can be explained by goals being generally closer to the origin with ChronoGEM than SMM or RND (c.f. \Cref{fig:helicopter}). Full results of the cross-validation with regards to varying $\epsilon$ are visualized in the Appendix, Figure~\ref{fig:cross_training}.
\\
We can better quantify the global results by collecting all the areas under the curve (AUC), and weighting them proportionally to the exponential of the evaluation goals' entropy. In effect, if a goal-set is very diverse, goals therein are more diverse and therefore more interesting to achieve. Conversely, if a goal-set has low entropy, it may simply contain the same couple of goals that are hard to achieve, but have low value in terms of learning a generalist controller. The exponential of the entropy quantifies the number of states in the distribution. We call this metric Entropy Weighted Goal Achievement (EWGA): 

$\text{EWGA}(\text{method}) =
\frac{\sum \limits_{s \in \text{eval sets}} e^{\text{entropy}(s)}* \text{AUC}(\text{method on }s)}{\sum \limits_{s \in \text{eval sets}} e^{\text{entropy}(s)}}$

\begin{figure*}
    \centering
    \includegraphics[width=\linewidth]{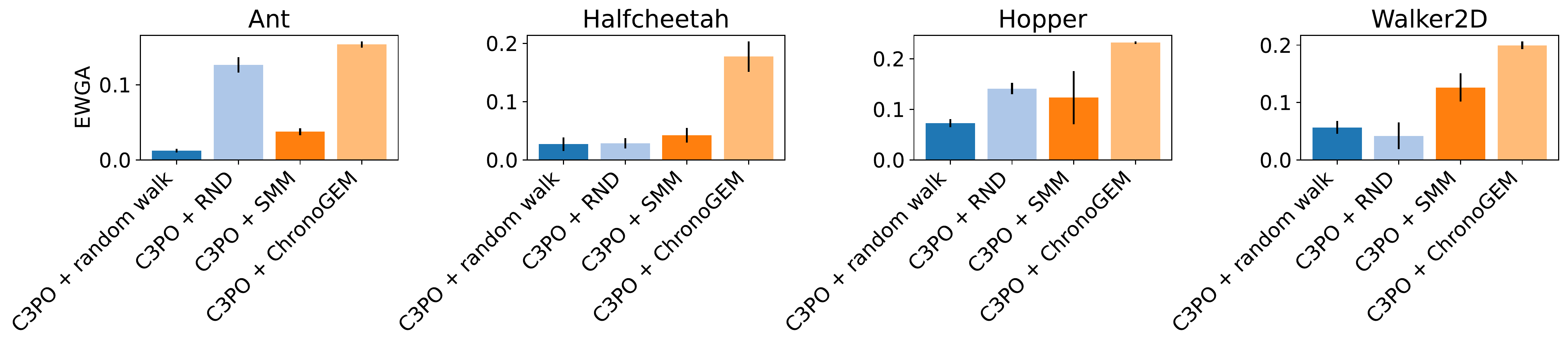}
    \vspace{-20pt}
    \caption{Entropy Weighted Goal-Achievement (EWGA). This estimates the ability of a policy to achieve goal sets that better covers the space (for example, a policy like C3PO that reaches a larger variety of states has an higher EWGA than a policy like SAC trained on the Random Walk, that only reaches states that are close to the origin).}
    \label{fig:ewga}
    \vspace{-5pt}
\end{figure*}

The performance of each C3PO variant is detailed in \Cref{fig:ewga}, where we can see that C3PO trained on ChronoGEM data is significantly better across all four environments regarding entropy-weighted AUC. To answer our original question on the importance of data quality, we can see that on datasets with higher entropy as per \Cref{fig:entropy}, downstream C3PO policy performance is higher both in cross-validation (\Cref{fig:cross_training}) and in entropy-weighted AUC (\Cref{fig:ewga}). In particular, ChronoGEM-trained C3PO is generally the most robust on un-normalized cross-validaiton, and is significantly superior on entropy-reweighted AUC.

\paragraph{Effects of Data Quantity.}
\label{massive_goal_training}
\label{sec:exp_general_quantity}
\begin{figure}[ht]
    \centering
    \includegraphics[width=0.8\linewidth]{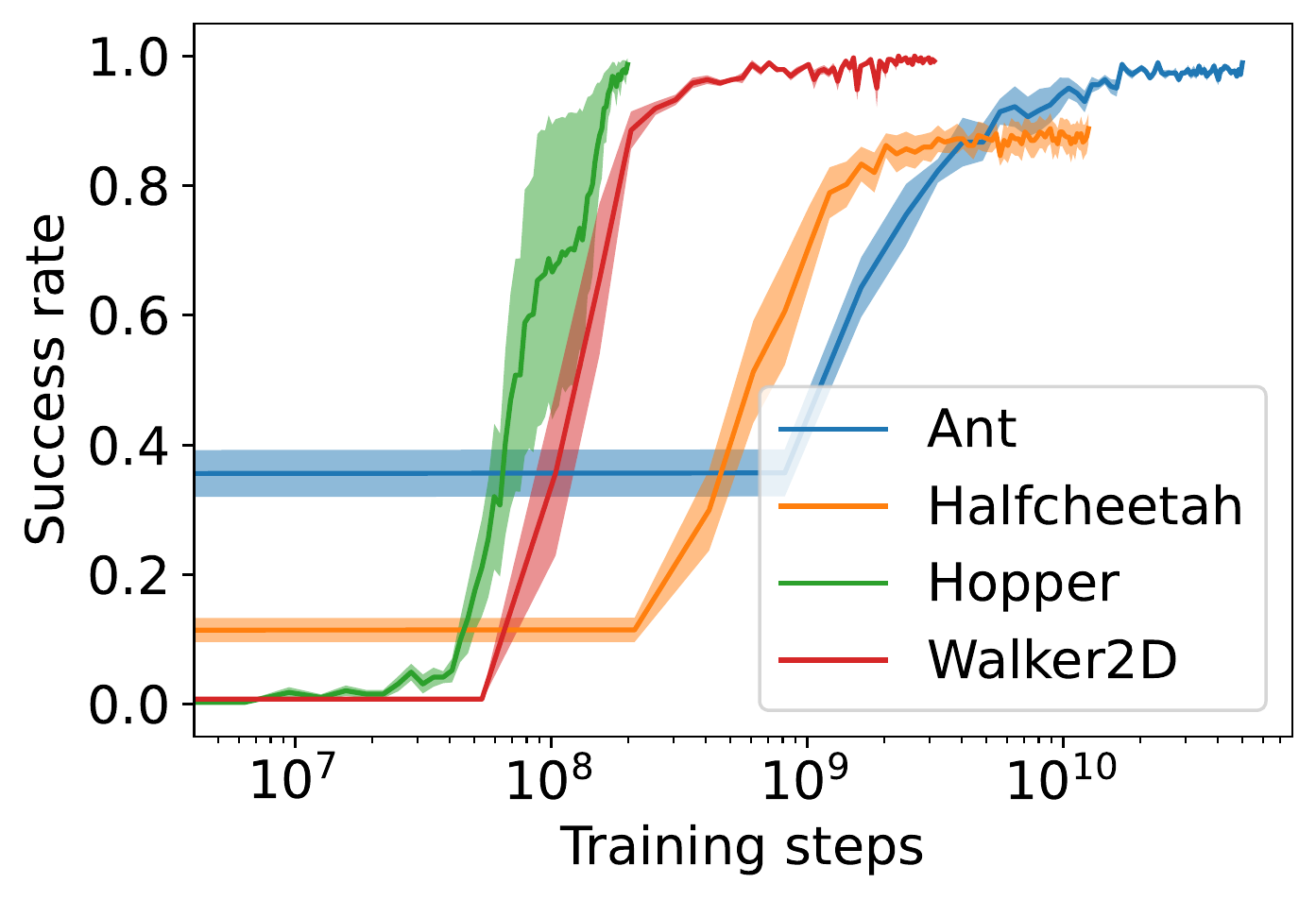}
    \vspace{-10pt}
    \caption{This plots illustrates C3PO-ChronoGEM training curves for each of the four environments used.  We can observe that although training can converge quickly on some environments, other require significant amounts of interaction.  Final convergence on Ant for example takes upwards of $10^{10}$ steps.}
    \label{fig:data_quantity}
    \vspace{-5pt}
\end{figure}
We now want to investigate the importance of large amounts of experiential data on the performance of a general goal-achievement policy. To do this, we will let C3PO train for many orders of magnitude more than what is generally used during continuous-control training regimes.  For questions of resource-efficiency, we restrict these experiments to the ChronoGEM datasets, as they seem to perform best for goal-achievement. For environments where we do not converge to a good controller earlier, we let them run for up to $30\times10^12$ steps.  Thanks to Brax's high parallelization and efficient infrastructure, it is possible to run such an experiment in a couple of days.  Although we did not use it in our earlier experiments for resource reasons, we also add the much more difficult Humanoid task to our set environments for this analysis, with a couple modifications described in \Cref{appendix:humanoid}.  In Figure XX we show the performance of each policy for a fixed $\epsilon$.  We can see that for certain environments such as Hopper and Walker2d train in tens of millions of steps.  One could be tempted to conclude that halfcheetah and ant have also hit their maximum performance at this stage.  However, we show that by continuing training for billions of steps, we can also achieve very good performance on these environments.  We illustrate some of the goal-achivement abilities of the policies able to achieve 90\% success at .25 tolerance are represented in Figure ~\ref{fig:best_frames} and in supplementary material\footnote{Please refer to \url{https://sites.google.com/view/chronogemc3po/home}}.  We believe this shows that not only is data quality important, but also providing significant amounts of experience is necessary for general goal-achievemnet policies to attain high levels of performance.

\begin{figure*}[ht]
    \centering
    \includegraphics[width=\linewidth]{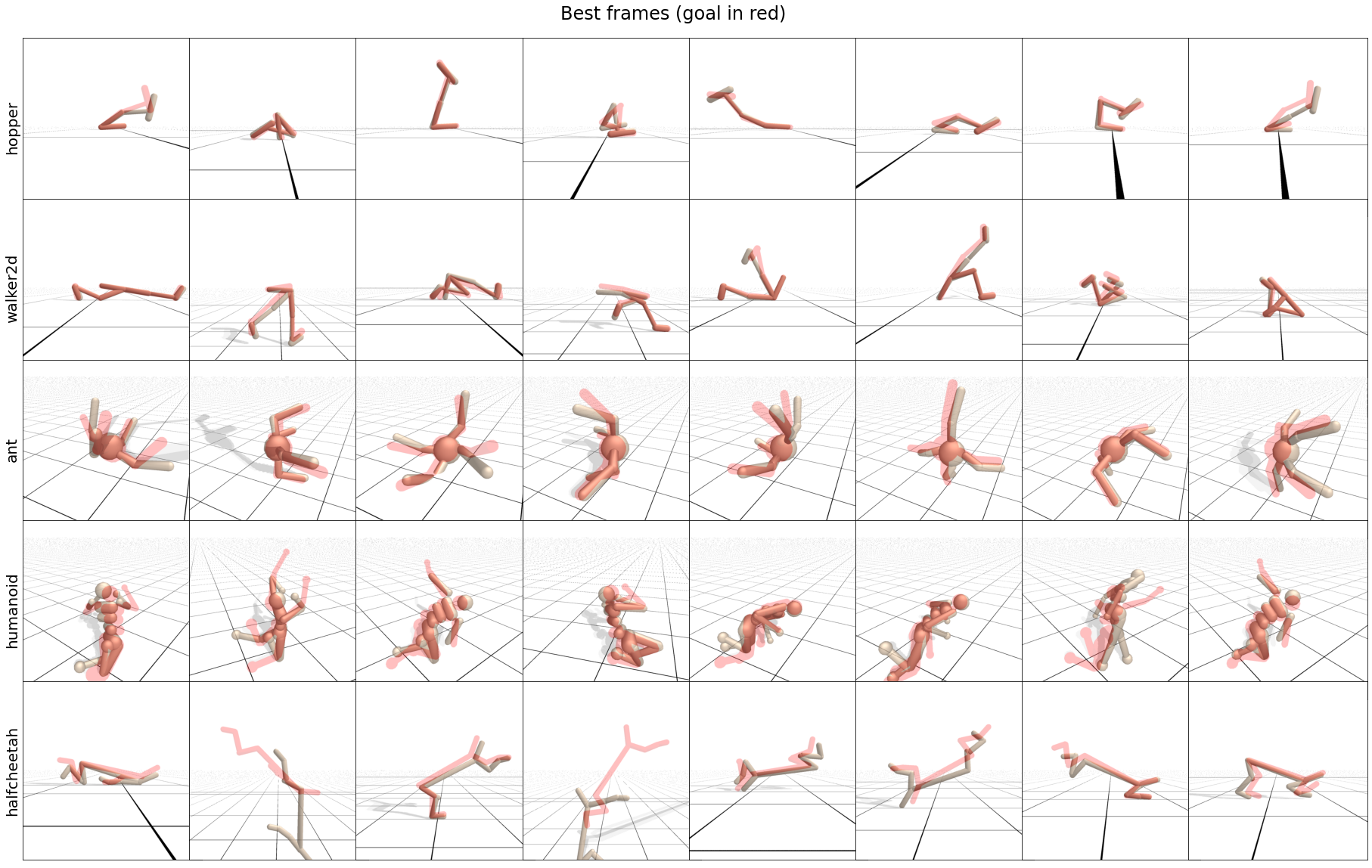}
    \vspace{-20pt}
    \caption{For each environment, we drew 8 goals from the ChronoGEM distribution and ran the policy trained on ChronoGEM goals. This figure represent the frame that is the closest from the goal for each episode, overlayed with the goal in red.}
    \vspace{-5pt}
    \label{fig:best_frames}
\end{figure*}

\subsection{Zero-shot imitation}
As an example task to illustrate the utility of a general goal-achievement policy we look at zero-shot imitation.  We can easily perform this by telling the C3PO policy to aim for successive target states from a pre-generated expert policy trajectory.
\\
we used a pre-trained C3PO policy to imitate an expert demonstration. Additional details on the implementation are in \Cref{appendix:zeroshot}. We define two types of task, one where a policy is trained from scratch on the base environment reward, which involves moving forward at a certain speed, and another where we take trajectories induced by ChronoGEM exploration (ChronoGEM has the side effect of generating original and energetic behaviours like cartwheels or backflips).   
For imitation of the base environment reward policies, we can evaluate the effective reward of the imitation policy. Results suggest that zero-shot imitation based on C3PO reaching demonstrator's states can achieve around 50\% of the expert score. For Ant, Halfcheetah, Hopper and Walker2d, the imitating agent successfully walked in the rewarding direction. Zero-shot imitation failed for humanoid, that could only manage to maintain a standing position but did not walk in the right direction. Full results are reported in in \Cref{tab:zero_shot}.
\\
Imitation of ChronoGEM trajectories was easier in the sense that all the states where by construction in-domain for the policy. Because it is difficult to quantitatively measure this performance in an interpretable way, we joined the resulting videos (both demonstration and imitation) in the supplementary material\footnote{Please refer to \url{https://sites.google.com/view/chronogemc3po/home}}. These videos show that Hopper and Walker almost perfectly imitate the expert, while Ant and Halfcheetah only ``approximate'' the demonstration (eg, halfcheetah realizing one front flip while the expert did two).  

\begin{table}
    \centering
    \begin{tabular}{|c|c|c|}
        \hline
        Environment & Expert & Zero-shot imitation  \\
        \hline
        Ant & 2281.45 & 1083.48 $\pm$ 317.42 \\
        \hline
        HalfCheetah & 1092.65 & 984.32 $\pm$ 112.49\\
        \hline
        Hopper & 676.58 & 214.95 $\pm$ 138.88\\
        \hline
        Walker2d & 965.53 & 562.26 $\pm$ 51.01\\
        \hline
        Humanoid & 2670.59 & 588.93 $\pm$ 505.45\\
        \hline
    \end{tabular}
    \caption{Zero-shot imitation of downstream tasks, based on a single expert demonstration of length 300. Imitation results are averaged over 9 different values (3 ChronoGEM seeds for C3PO $\times$ 3 environment seeds).}
    \label{tab:zero_shot}
\end{table}

\section{Related works}
\label{sec:related}
This work is situated between various fields.  Although effectively a goal-conditioned policy optimization algorithm, C3PO is enabled by the ChronoGEM exploration algorithm.  We will first look at similar exploration methods and then consider various goal-conditioned learning setups.

\subsection{Bonus-based exploration}
Although generally not concerned with goal-conditioned RL, there is a large family of exploration methods that are manifest as reward bonuses, with the intent of training a policy faster, or to be more efficient.
One family of approaches uses state-visitation counts that can be approximate to create an associated bonus for rarely-visited states~\citep{bellemare2016unifying, ostrovski2017count}. 
Prediction-error bonuses use the residual error on predictions of future states as a reward signal to approximate novel states, this includes methods such as RND~\citep{burda2018exploration} which leverages the prediction error of random projections of future states, or SPR~\citep{schwarzer2020data} and BYOL-Explore~\citep{guo2022byol}, which make use of the self-prediction error of the network with a frozen version of itself.
Model-based methods often optimise for next-state novelty, either by looking at the KL divergence between sampled states and likely states, such as in  VIME~\citep{houthooft2016vime} or by explicitly driving towards states with high model ensemble disagreement such as in Plan2Explore~\citep{sekar2020planning}.
RIDE~\citep{raileanu2020ride} and NGU~\citep{badia2020never} use episodic memory in which the bonus reflects the variety of different states covered in a single trajectory.

\subsection{Diffusion-based exploration}\label{rw:diffusion}
ChronoGEM is based on a tree-structured diffusion, that makes a selection of states, randomly explores from these states and then reselect states, and so on. Go-Explore~\citep{ecoffet2019go} uses a similar approach, by running a random policy for some steps, selecting a set of `interesting' states, and then branching from these.  ChronoGEM does not require returning to said states, and only requires one step of random actions at each iteration. An idealized ChronoGEM  with perfect density estimation is additionally provably approximating a uniform distribution over achievable goals as the number of sampled states is large enough, and it does not require any additive prior regarding state importance. 
Another close work also using a diffusion approach is UPSIDE~\citep{kamienny2021direct}. It finds a set of nodes along with a set of policies that connect any node to the closest ones, looks for new nodes by random exploration from the existing ones, and removes  unnecessary nodes that are reached by the less discriminant policies. 
UPSIDE converges to a network of nodes that efficiently covers the state space. However, by construction UPSIDE does not produce uniform coverage, but a set of policies that reach different regions of the space. Using UPSIDE to obtain a similar quantity of states uniformly distributed than in ChronoGEM, we would have to train UPSIDE with $2^{17}$ policies, which is not  computationally feasible.

\subsection{Entropy maximisation}
Some exploration algorithms, such as ChronoGEM, are constructed in order to maximize the entropy of the state visitation distribution. Most of them, however, focus on the distribution induced by the whole history buffer (instead of the just $T$-th states of episodes in ChronoGEM), generally based on the behavior of a trained policy. This is the case of MaxEnt~\citep{hazan2019provably}, GEM~\citep{guo2021geometric}, SMM~\citep{lee2019efficient} and CURL~\citep{geist2021concave}.
In APT~\citep{liu2021behavior}, instead of using a density model to estimate the entropy, they use a non-parametric approach based on the distance with the K nearest neighbors in a latent representation of the state space. APS~\citep{liu2021aps} combines APT's entropy bonus with an estimation of the cross-entropy based on successor features to maximize the mutual information $I(w; s)$ between a latent skill representations $w$ and states.

\subsection{Goal-Conditioned Reinforcement Learning}
Goal-conditioned RL~\citep{kaelbling1993learning, schaul2015universal} is the general setup of learning a goal-conditioned policy instead of a specialized policy. We are particularly interested in goal-based setups where there is no \textit{a-priori} reward function.  Although well known works such as HER~\citep{andrychowicz2017hindsight} demonstrate methods for learning goal-conditioned policies with minimal explicit exploration, more recent works~\citep{pitis2020maximum, openai2021asymmetric,mendonca2021discovering} demonstrate the importance of having a good curriculum of goals to train from. MEGA~\citep{pitis2020maximum} extends HER-style relabeling and answers the exploration problem by iteratively sampling goals according to a learnt density model of previous goals. ABC~\citep{openai2021asymmetric} demonstrates the importance of an adversarial curriculum for learning more complex goal-conditioned tasks, but is concentrated on specific tasks instead of arbitrary goal achievement. LEXA~\citep{mendonca2021discovering} builds on Plan2Explore~\citep{sekar2020planning}, and demonstrates the importance both of a good exploration mechanism, as well as the use of significant amounts of (imagined) data for learning an arbitrary goal-achievement policy.   DIAYN~\citep{eysenbach2018diversity} uses a two-part mechanism that encourages the agent to explore novel areas for a given latent goal, while at the same time learning a goal embeddings for different areas of the state space. 
While some of the above methods consider notions of density for exploration~\citep{eysenbach2018diversity}, C3PO uses a more principled exploration mechanism, and is particularly interested in collecting a large and uniformly distributed dataset of reachable states to train a goal-conditioned policy.

\section{Conclusion}

In this paper we looked at generating massive amounts of data on MDPs with no prior task information, as well as the effects of data quantity and quality on training a general policy.  When compared to similar methods, we show that our proposed exploration mechanism ChronoGEM is capable of generating massive amounts of useful data.  We also show that when training a downstream general controller policy with C3PO, the quality of ChronoGEM data for goal states is superior to similar methods, and that letting the policy consume massive amounts of experience data is fundamental to achieving high-quality general controllers.  We believe these insights suggest that general controllers are achievable, but will require both performant exploration such as that generated by ChronoGEM and potentially massive amounts of data to be able to achieve high performance.

\clearpage
\bibliographystyle{apalike}
\bibliography{c3po}

\clearpage
\appendix
\onecolumn

\section{Uniform sub-sampling}\label{appendix:uniform_sampling}
Let $f_X$ be the density of a distribution with domain $S\subset\R^d$ ($x\in S \Leftrightarrow f_X(x)>0$), and $X_1\hdots X_n\sim f_X$ iid. We assume that $S$ is bounded.
We define the sub-sampling $Y_n$ such that $P(Y_n=X_i\vert X_1\dots X_n)\propto \frac{1}{f_X(X_i)}$, and $Y\sim \uc_S$ a random vector following the uniform distribution over $S$.
\begin{theorem}
$Y_n$ conditioned to $X_1\dots X_n\sim f_X$ iid converges in distribution to $Y$ when n goes to infinity.
\label{thm:chronogem}
\end{theorem}
\begin{proof}
let $A$ be any subset of $S$, and $\mu(.)$ the Lebesgue measure over $S$. Given any sampling $X_1\dots X_n\sim f_X$ iid, we have:
\begin{align}
    P(Y_n\in A\vert X_1\dots X_n) =  \frac{\sum_{X_i\in A}\frac{1}{f_X(X_i)}}{\sum_{j=1}^n\frac{1}{f_X(X_j)}}
    = \frac{\frac{1}{n}\sum_{X_i\in A}\frac{1}{f_X(X_i)}}{\frac{1}{n}\sum_{j=1}^n\frac{1}{f_X(X_j)}} \rightarrow \frac{\E_X[\un_{X\in A}\frac{1}{f_X(X)}]}{\E_X[\frac{1}{f_X(X)}]} = \frac{\mu(A)}{\mu(S)}.
\end{align}
\end{proof}

\section{Density Estimator Model Selection}\label{appendix:density}
We implemented 7 candidate models, including Gaussian models (Multivariate, Mixture), autoregressive networks (RNade~\citep{uria2013rnade}, Made~\citep{germain2015made}), and normalizing flows (real-NVP~\citep{dinh2016density}, Maf~\citep{papamakarios2017masked}, NSF~\citep{durkan2019neural}).  We decided to compare them on two continuous control task from the Brax environments~\citep{brax2021github}: Ant and HalfCheetah.

We proceeded in 5 steps.

\subsection{Policy pre-training}
We first pre-trained policies to solve all the different tasks, using 4 different RL algorithms: a uniform random walk, PPO, SAC and Evolution Strategies. 

\subsection{Density model training}
For each pre-trained policy, we trained each density model with various hyperparameters configurations to maximize the log-density of the states visitation. Table~\ref{tab:hypers} reports, for each model, all the values we tried for each hyperparameter of the model, and the different configurations we used are all combinations of these values. For each model, the training consisted in 100 epochs, each one containing a training phase interacting with 128$\times$1000 environment transitions and an evaluation phase based on the log-likelihood score averaged over 128$\times$1000 visited states. Every training was run over 5 different seeds and resulting scores were averaged across these seeds.

\begin{table}[ht]
    \centering
    \begin{tabular}{|l|l|l|l|}
        \hline
        \multicolumn{2}{|c|}{\textbf{Gaussian}} &  \multicolumn{2}{|c|}{\textbf{Made}}\\
        \hline
        Learning rate & 5e-4, \textbf{1e-4}, 1e-3 & Learning rate & \textbf{5e-4}, 1e-4, 5e-4\\
        \hline
        Batch size &  32, 64, \textbf{128} & Batch size &  \textbf{32}, 64, 128\\
        \hline
        \multicolumn{2}{|c|}{\textbf{Mixture of Gaussian}} & Num. masks & 1, \textbf{2}, 4\\
        \hline
        Learning rate & 5e-4, 1e-4, \textbf{1e-3} & Num. Mixtures & 1, 4, \textbf{8}\\
        \hline
        Batch size &  32, 64, \textbf{128} & Num. hidden layers & \textbf{2}\\
        \hline
        Num. mixtures &  10, 20, \textbf{30} & Hidden layers dim. & 5, 10, \textbf{50}\\
        \hline
        \multicolumn{2}{|c|}{\textbf{RNade}} & \multicolumn{2}{|c|}{\textbf{real-NVP, Maf, NSF} }\\
        \hline
        Learning rate & 5e-4, \textbf{1e-4}, 5e-4 & Learning rate & \textbf{5e-4}, 1e-4, 5e-4\\
        \hline
        Batch size &  \textbf{32}, 64, 128 & Batch size &  32, 64, \textbf{128} (Real-NVP: \textbf{64})\\
        \hline
        Num. mixtures & 1, 10, \textbf{20}, 30& Num. layers &  4, 6, \textbf{8} (MAF: \textbf{6})\\
        \hline
        Hidden layer dim. &  10, 20, \textbf{50} & hidden layers per layers &  \textbf{500$\times$500}\\
        \hline
    \end{tabular}
    \caption{Explored hyperparameters. Bolded values are correspond to the best configuration for each model, based on AUC criteria.
    For normalizing flows, we found the same best configurations, except a batch size of 64 for Real-NVP (128 for MAF and NSF) and 6 layers for MAF (8 for Real-NVP and NSF).}
    \label{tab:hypers}
\end{table}

\subsection{Configuration selection}\label{conf_select}
For each model, we selected the best configuration across all environments and all RL algorithm, based on AUC. 
For each model, to compare the performance of configuration across various environment and algorithm, we normalized the score w.r.t. the max and min performances for each environment and RL algorithm couple. This procedure is described in Algo~\ref{algo:hps}.
Bold values in Table~\ref{tab:hypers} represent the hyper from the best configuration we found for each model.

\begin{algorithm}[ht]
    \caption{Hyper parameters selection for a given model.}
    \label{algo:hps}
    
\textbf{Input}: $\{\text{rl}_i\}$ a set of RL algorithms, $\{\text{env}_j\}$ a set of environment and $\{\text{conf}_k\}$ a set of hyperparams configuration. Score function $S(\text{rl}_i, \text{env}_j, \text{conf}_k)\in \R$, for ex AUC.

\hspace{5mm}

\textcolor{gray}{Normalize configurations scores for each RL algorithm and environment:}

\textbf{For} $\text{rl}_i$, $\text{env}_j \in \{\text{rl}_i\} \times \{\text{env}_j\}$:

    \hspace{5mm} $M(\text{rl}_i, \text{env}_j) = \max_k S(\text{rl}_i, \text{env}_j, \text{conf}_k)$
    
    \hspace{5mm} $m(\text{rl}_i, \text{env}_j) = \min_k S(\text{rl}_i, \text{env}_j, \text{conf}_k)$
    
    \hspace{5mm} \textbf{For} $\text{conf}_k \in \{\text{conf}_k\}$:
    
    \hspace{10mm} $\bar{S}(\text{rl}_i, \text{env}_j, \text{conf}_k) = \frac{S(\text{rl}_i, \text{env}_j, \text{conf}_k) - m}{M - m}$
    
\hspace{5mm}

\textcolor{gray}{Sum normalized scores across different RL algorithms and environments:}

\textbf{For} $\text{conf}_k \in \{\text{conf}_k\}$:
\hspace{5mm} $\mathcal{F}(\text{conf}_k) = \sum_{i, j}\bar{S}\text{rl}_i, \text{env}_j, \text{conf}_k)$

\hspace{5mm}

\textcolor{gray}{Return best configuration according to sum of normalized scores:}

\textbf{Return}: $\argmax_{k}\mathcal{F}(\text{conf}_k)$

\end{algorithm}

\subsection{Hyperparameters effects}
We studied the effect of hyper parameters, based on two approaches. In the first one, given each value of each single hyper we selected the best configuration for the rest of hyper parameters and looked at the resulting score. In the second one, given a selected best configuration (from the step described above in~\ref{conf_select}) we replaced the value of one hyper by another one and looked how it deteriorated the score.
However, we found no surprising effect: globally, the larger the models, the higher the score. Parameters that did not affect the size of the model (for ex the batch size) had no significant effect, especially on normalizing flows.

\subsection{Model comparison}
\begin{figure}[ht]
    \centering
    \includegraphics[width=\linewidth]{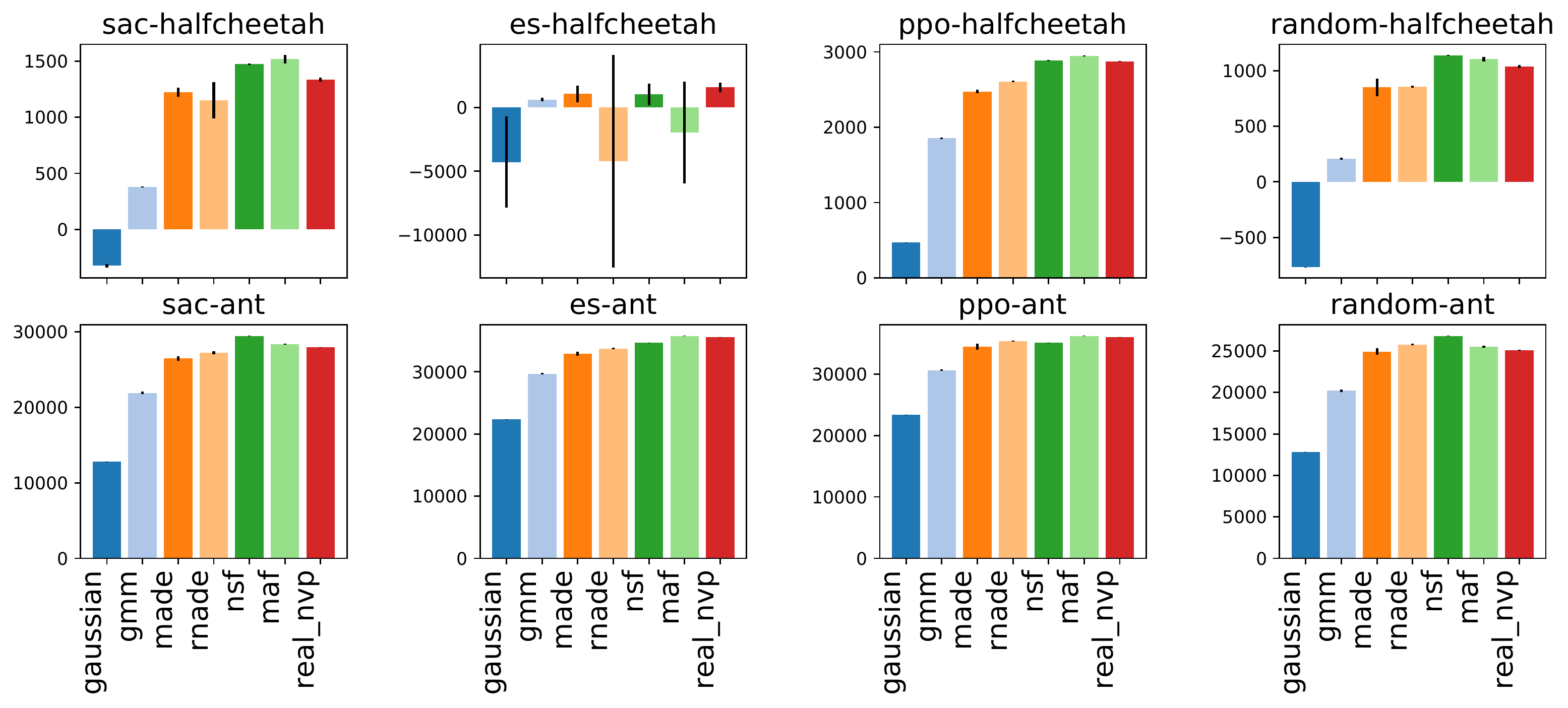}
    \caption{Model comparison based on AUC with selected best hyper configuration across all RL algorithms and environments, averaged over 5 seeds.}
    \label{fig:auc}
\end{figure}

Finally, we compared the different models together on the different environments and RL algorithms, when run with the selected best configuration (from step~\ref{conf_select}). Visually Gaussian and Mixture of Gaussians are less efficient than autoregressive models, which are less efficient than normalizing flows. Among normalizing flows, we found that NSF had the best average score across all RL algorithm and environments, while being the less sensible to the variations of hyper parameters. Figure~\ref{fig:auc} reports bar plots comparison of all model's AUC scores given their best hyper configurations on all the different RL algorithms and environments, averaged over 5 seeds.

\section{Space visitation of 2D environments}\label{appendix:visitation}

\begin{figure}
\centering  
\begin{subfigure}
     \centering
     \includegraphics[width=\textwidth]{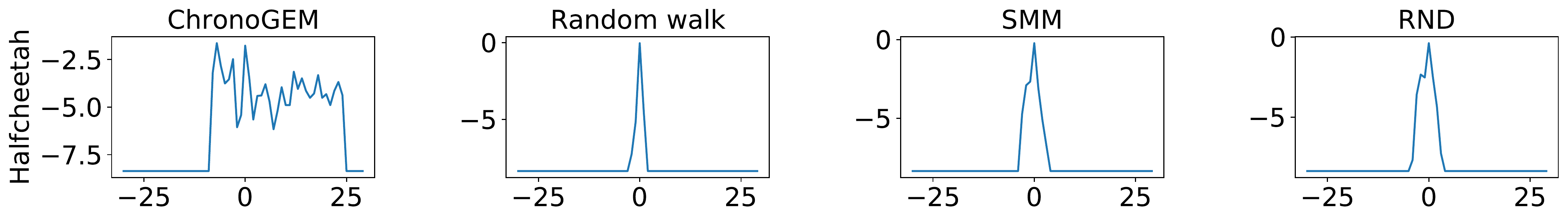}
\end{subfigure}
\begin{subfigure}
     \centering
     \includegraphics[width=\textwidth]{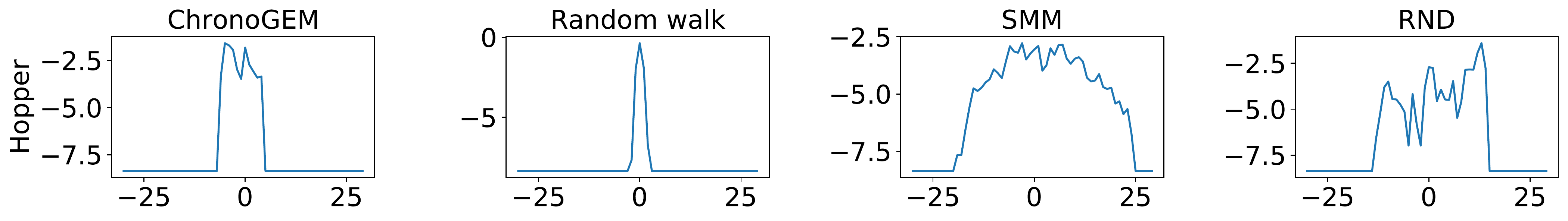}
\end{subfigure}
\begin{subfigure}
     \centering
     \includegraphics[width=\textwidth]{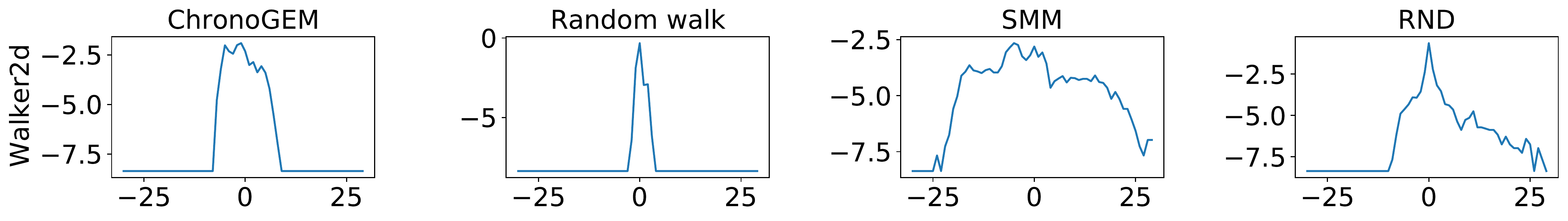}
\end{subfigure}
\caption{Log-frequencies of discretised X-axis visitations in the 2-dimensional environments (Hopper, Walker2d and Halfcheetah). In Hopper and Halfcheetah, SMM and RND visited a larger scope of spatial positions, but actually neglected to explore the possible poses, while ChronoGEM had a more uniform behaviour and well balanced both poses and positions exploration. In Halfcheetah, only ChronoGEM was able to run a decent exploration under the action reduction with a low multiplier.}
\label{fig:visits}
\end{figure}

See Figure~\ref{fig:visits}.

\subsubsection{Resettable states assumption}
\label{app:resettable}
Similarly to other diffusion-based algorithms (see related works \ref{rw:diffusion}), ChronoGEM needs to explore many actions from a single given state.  Although this is an unrealistic assumption for applications involving real-world systems, we argue that there are multiple scenarios where this is an acceptable assumption.  To begin with, many tasks exist only as software: computer games~\citep{berner2019dota, roydirectbehavior}, physics simulations~\citep{jumper2021highly, schmidt2019recent}, software-aided design~\citep{mirhoseini2021graph} or even arbitrary programs~\citep{schwartz2019autonomous, bottinger2018deep} are all important tasks that can be arbitrarily reset and parallelized.  Secondly, sim2real~\citep{tan2018sim, smith2022legged} approaches allow for a policy trained on simulation to be transferred to real systems.  In the case where a particular robot embodiment such as a quadruped walker might be used for a large number of task, it would make sense to invest the time in establishing a high-fidelity simulator and a sim2real pipeline to train generalist controllers such as those generated by C3PO.
\section{Exploration methods to reaching states from other exploration methods}\label{appendix:cross_expe}

\begin{figure*}[ht]
    \centering
    \includegraphics[width=\linewidth]{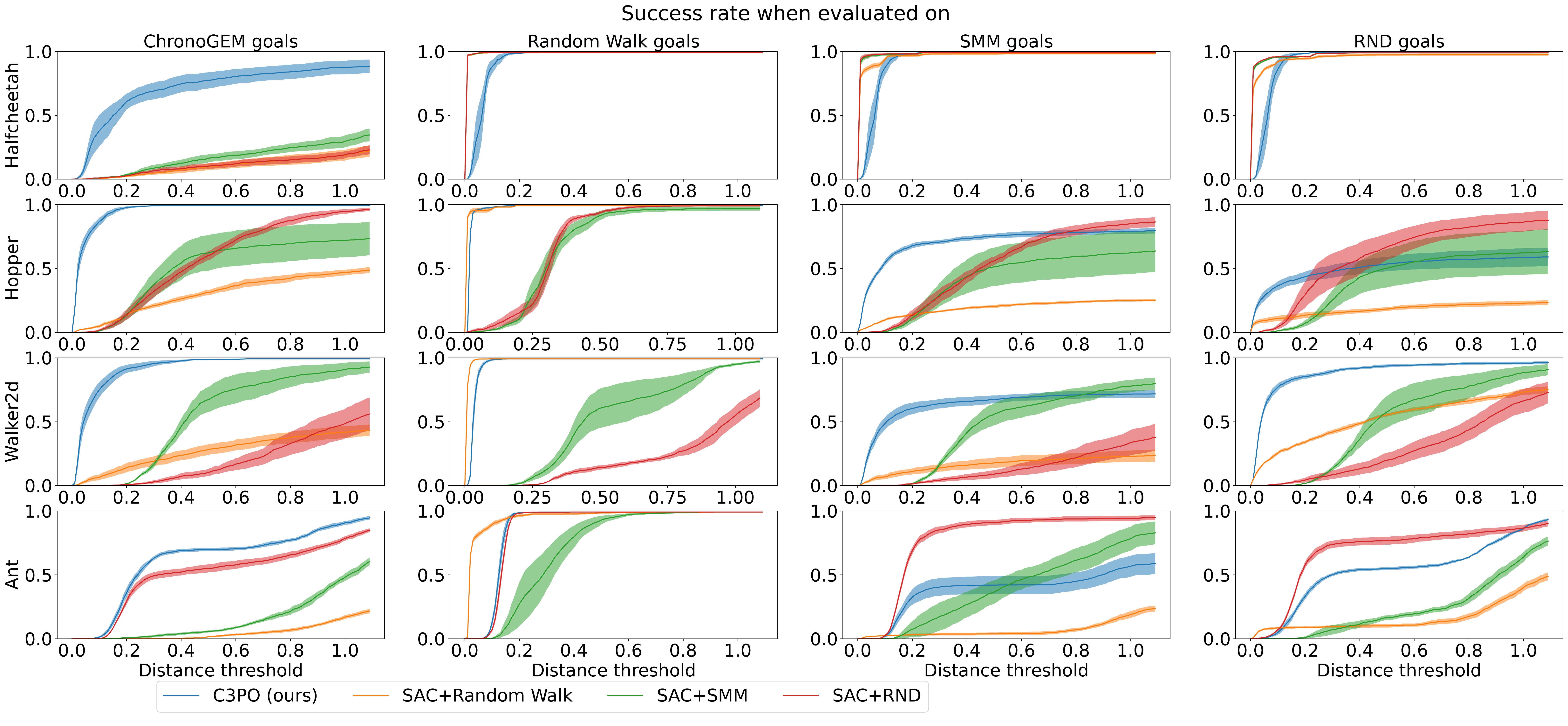}
    \caption{For each environment (lines) and each set of evaluation goals (columns), success rates as a function of distance thresholds obtained by SAC when trained on the different sets of training goals (\textcolor{blue}{ChronoGEM}, \textcolor{orange}{Random Walk}, \textcolor{teal}{SMM}, \textcolor{red}{RND}). Each exploration algorithm was run over 3 seeds to collect evaluation and training goals, and each SAC training was also run over 3 seeds, so the resulting curves are actually averaging 9 different values.}
    \label{fig:cross_training}
\end{figure*}

See Figure~\ref{fig:cross_training}.

\section{Method Hyper-Parameters}\label{appendix:details}
See \Cref{tab:chrono-hypers} and \Cref{tab:c3po-hypers} for C3PO hypers.
\subsection{Specifics of ChronoGEM for Humanoid}
\label{appendix:humanoid}
By default, ChronoGEM would mostly explore positions where the humanoid is on the floor.
However, it was simple to modulate the algorithm to only explore uniformly in the space of state where the humanoid is standing. For example, on can associate zero weight to undesired states during the re-sampling step. Thus, we avoid states in which the torso drops below an altitude of .8 (the default failure condition).
ChronoGEM is modified to not draw states where the humanoid is too low. The goal-conditioned learner gets a high penalty for going too low as well.  This demonstrates that when in posession of a prior, we can leverage it to steer the exploration and policy learning.

\subsection{Specifics for Zero-Shot Imitation}
\label{appendix:zeroshot}
Since reaching a target state is never immediate and requires at least a few action steps, we sub-sampled the expert trajectory to take one state every $n$ states as a target for the imitating agent. Harder tasks would require higher values of $n$, making the imitation task  less strict. 
\\
Since the agent was trained on ChronoGEM targets that are reachable in 128 steps, we considered relatively small episodes of 300 steps. Otherwise, as the imitator is moving slower than the expert, at some point the distance between the imitator's state and the target expert's states goes out of C3PO's domain of knowledge.
For downstream tasks, we could directly evaluate the quality of imitations by looking at the environment's returns. We used a single expert demonstration and averaged results over 9 different values (3 ChronoGEM seeds for C3PO $\times$ 3 environment seeds). Full results of policy-imitation are available in \Cref{tab:zero_shot}.

\begin{table}[h]
    \centering
    \begin{tabular}{|l|l|}
        \hline
        \multicolumn{2}{|c|}{\textbf{ChronoGEM Hypers}}\\
        \hline
        buffer size & $2^{17}$\\
        \hline
        time horizon & 128\\
        \hline
        branching factor & 4\\
        \hline
        NSF batch size & $2^{11}$\\
        \hline
        NSF learning rate & 3e-5\\
        \hline
        NSF hidden layers size & [512, 512]\\
        \hline
        NSF number of hidden layers & 8\\
        \hline
        NSF training epochs per step & 1\\
        \hline
    \end{tabular}
    \caption{ChronoGEM Hyperparameters}
    \label{tab:chrono-hypers}
\end{table}

\begin{table}[h]
    \centering
    \begin{tabular}{|l|l|}
        \hline
        \multicolumn{2}{|c|}{\textbf{SAC Hypers}}\\
        \hline
        normalize observations & True\\
        \hline
        reward scaling & 1.\\
        \hline
        number of actors & $2^{10}$\\
        \hline
        batch size & 1024\\
        \hline
        discounting & .98\\
        \hline
        learning rate & 3e-5\\
        \hline
        min replay size & $2^{13}$\\
        \hline
        max replay size & $2^{20}$\\
        \hline
        epsilon update threshold & .9\\
        \hline
        epsilon update multiplier & .99\\
        \hline
        gradient updates per actor episode & 32 * episode length\\
        \hline
        networks size & (1024) * 4\\
        \hline
    \end{tabular}
    \caption{C3PO-SAC Hyperparameters}
    \label{tab:c3po-hypers}
\end{table}

\section{Example States Generated by Each Method}
\begin{figure}
\includegraphics[width=\linewidth]{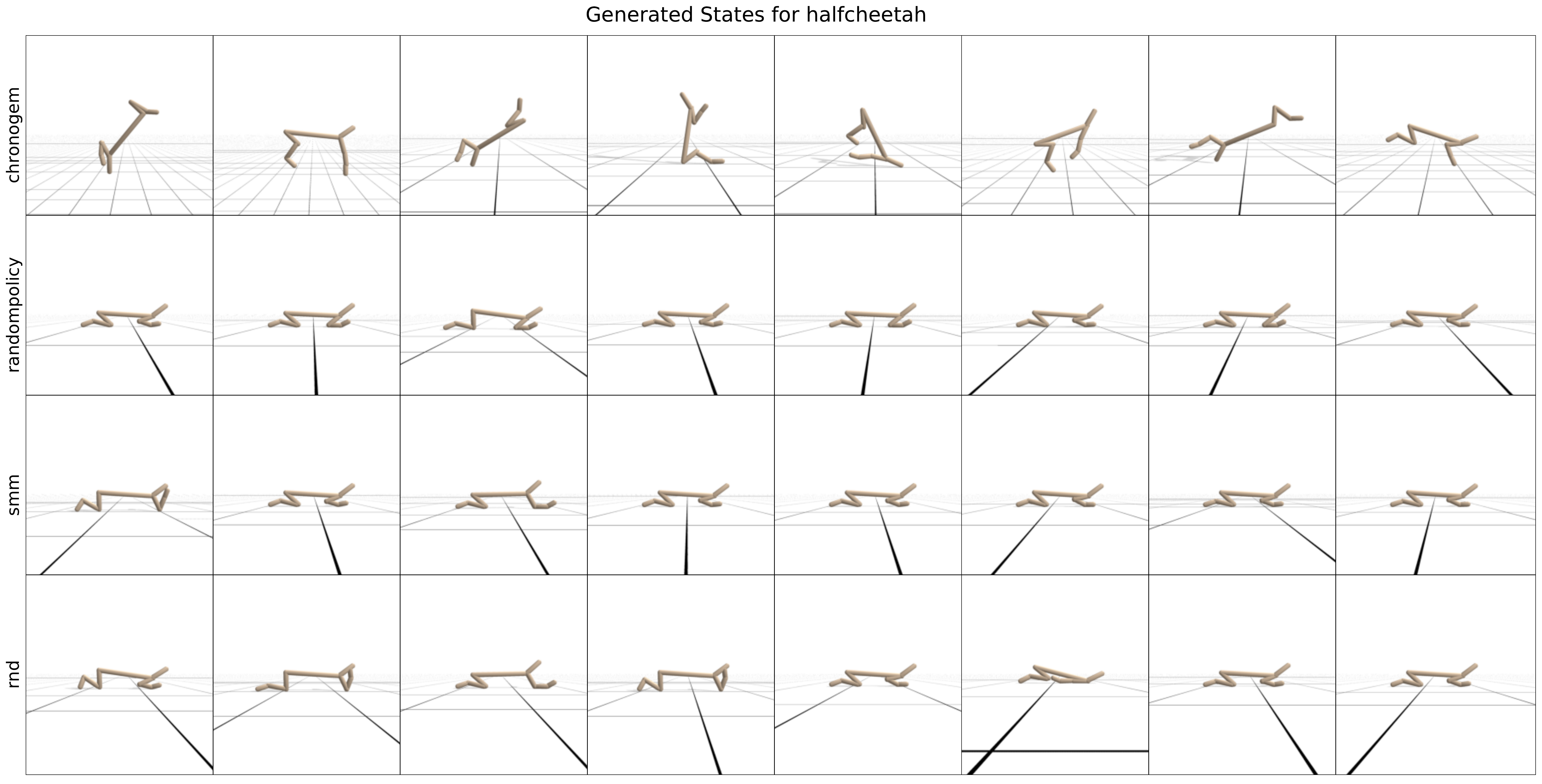} \\
\includegraphics[width=\linewidth]{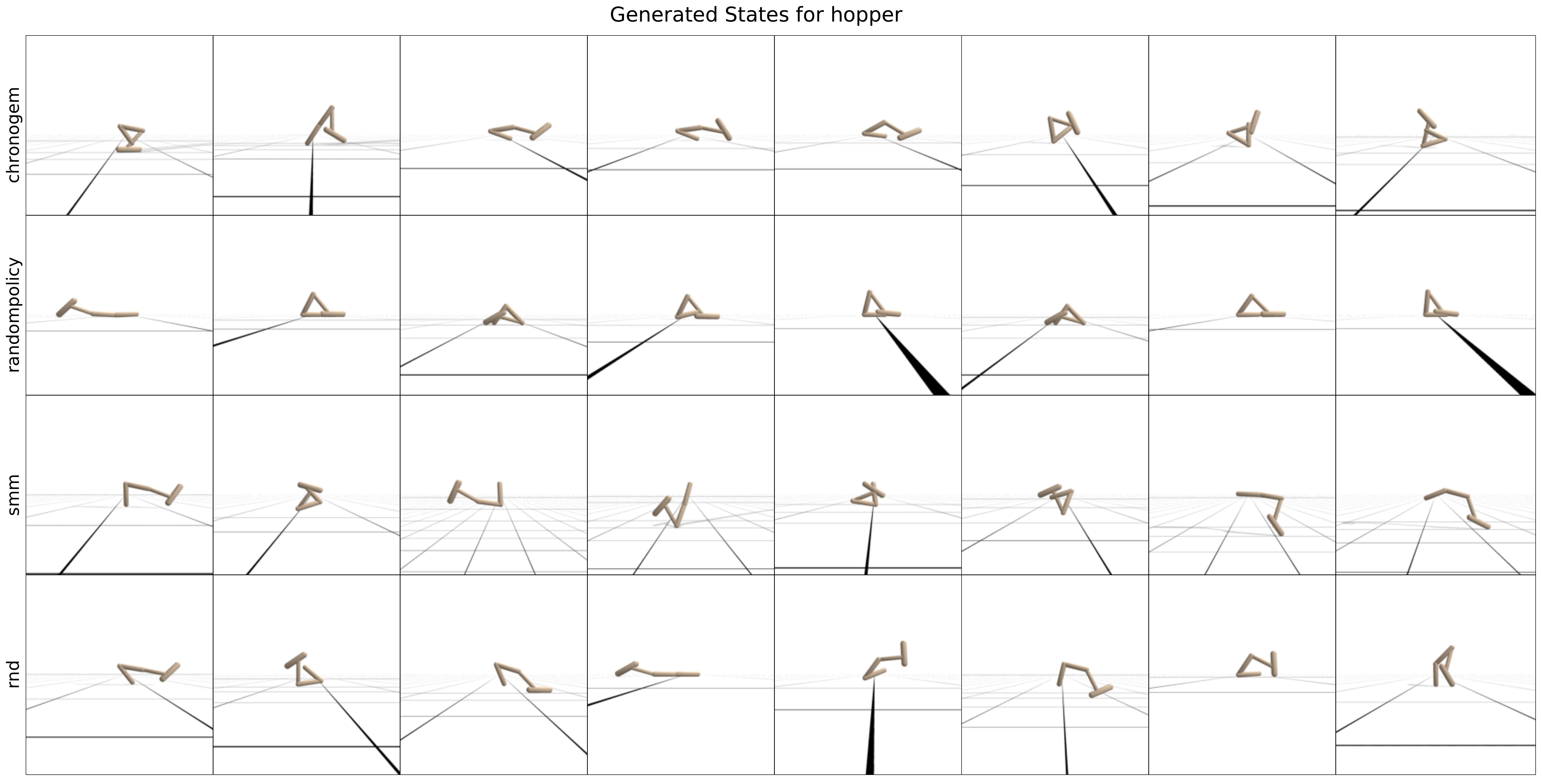}
\caption{Random samples of generated states from the various exploration methods investigated for the HalfCheetah and Hopper environments.  We can observe reduced variability for \texttt{randompolicy} states, as well as \texttt{smm} states in halfcheetah.  These correspond to the reduces overall performance of policies trained on these datasets for HalfCheetah and Ant.  Each visualized state is sampled directly from the dataset used for training goal-conditioned policies.}
\end{figure}
\begin{figure}
\includegraphics[width=\linewidth]{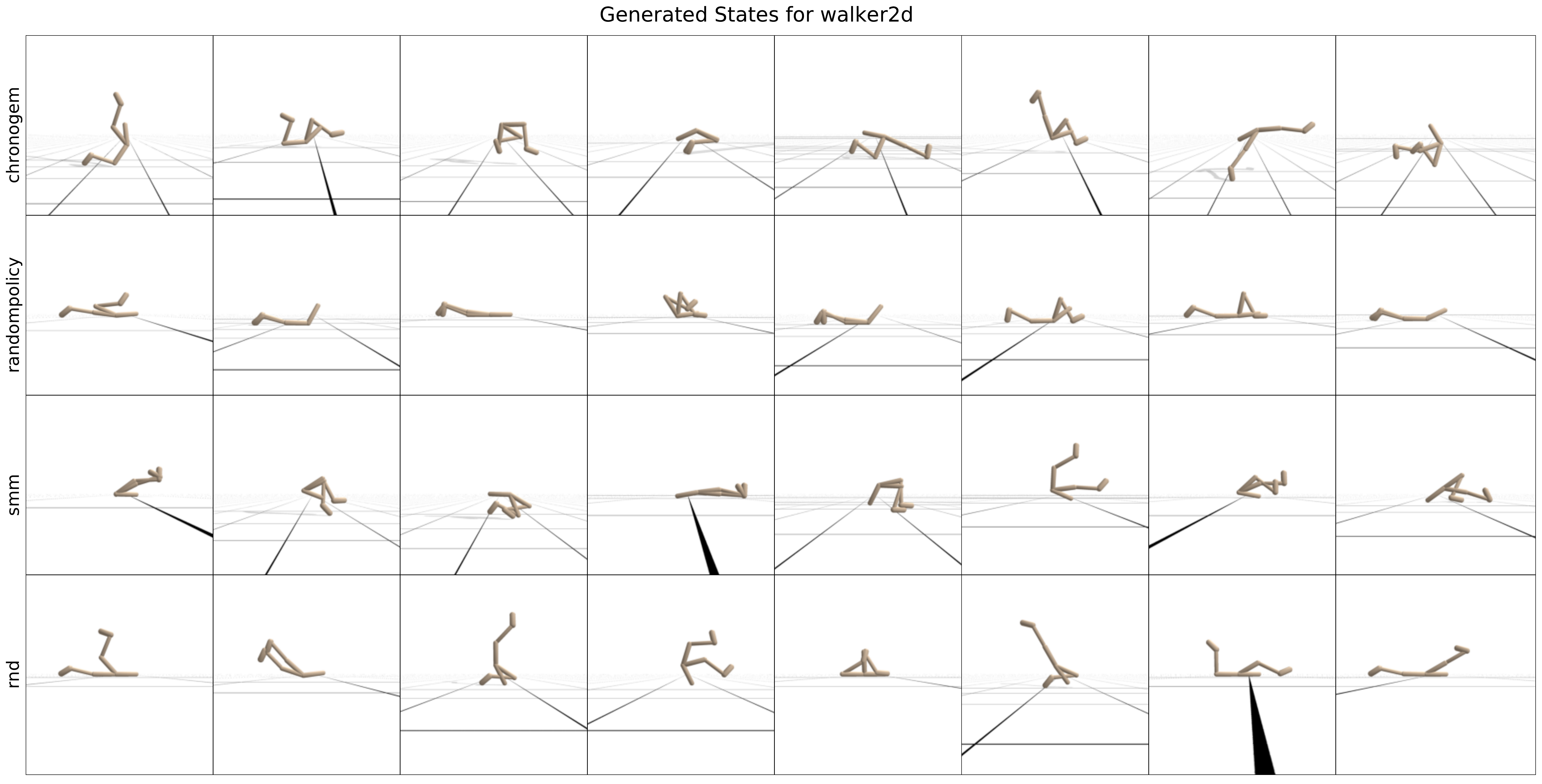} \\
\includegraphics[width=\linewidth]{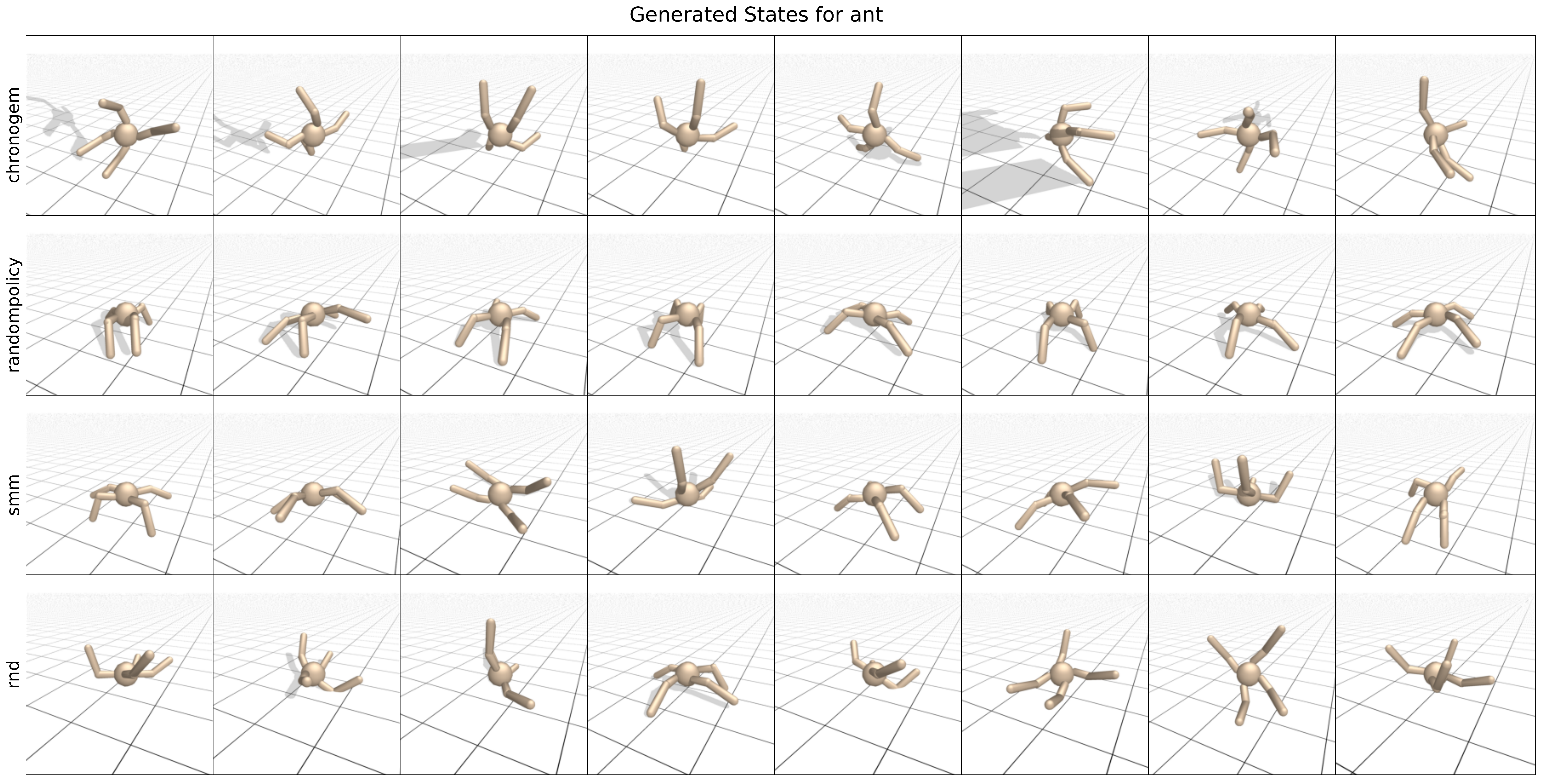}
\caption{Random samples of generated states from the various exploration methods investigated for the Walker2D and Ant environments.  As in the previous figure we can observe reduced variability for \texttt{randompolicy} states.  Each visualized state is sampled directly from the dataset used for training goal-conditioned policies.}
\end{figure}
\end{document}